\documentclass[sigconf]{acmart}
\usepackage{booktabs}
\usepackage{multirow}
\usepackage[normalem]{ulem}
\useunder{\uline}{\ul}{}
\usepackage{xcolor}
\usepackage{makecell}
\usepackage{tcolorbox}
\tcbuselibrary{breakable} 
\usepackage{pifont}
\usepackage{algorithmic}
\usepackage{algorithm}
\usepackage{enumitem}
\definecolor{myblue}{RGB}{150, 150, 230}
\AtBeginDocument{%
  }
    
\usepackage{enumitem}
\usepackage{booktabs}

\setcopyright{acmlicensed}
\copyrightyear{2018}
\acmYear{2018}
\acmDOI{XXXXXXX.XXXXXXX}
\acmConference[Conference acronym 'XX]{Make sure to enter the correct
  conference title from your rights confirmation email}{June 03--05,
  2018}{Woodstock, NY}
\acmISBN{978-1-4503-XXXX-X/2018/06}

\acmSubmissionID{123-A56-BU3}

\begin{document}

\title[
A Holistic Benchmark and Agent Framework for Complex Instruction-based Image Generation
]{
Draw ALL Your Imagine: A Holistic Benchmark and Agent Framework for Complex Instruction-based Image Generation
}

\author{Yucheng Zhou$^1$, ~Jiahao Yuan$^2$, ~Qianning Wang$^3$} 
\affiliation{
  \institution{$^1$University of Macau \country{China}, ~$^2$East China Normal University \country{China}, \\~$^3$Auckland University of Technology \country{New Zealand}}
}
\email{yucheng.zhou@connect.um.edu.mo}

\renewcommand{\shortauthors}{Zhou et al.}

\begin{abstract}
Recent advancements in text-to-image (T2I) generation have enabled models to produce high-quality images from textual descriptions. However, these models often struggle with complex instructions involving multiple objects, attributes, and spatial relationships. Existing benchmarks for evaluating T2I models primarily focus on general text-image alignment and fail to capture the nuanced requirements of complex, multi-faceted prompts.
Given this gap, we introduce LongBench-T2I, a comprehensive benchmark specifically designed to evaluate T2I models under complex instructions. LongBench-T2I consists of 500 intricately designed prompts spanning nine diverse visual evaluation dimensions, enabling a thorough assessment of a model's ability to follow complex instructions. 
Beyond benchmarking, we propose an agent framework (Plan2Gen) that facilitates complex instruction-driven image generation without requiring additional model training. This framework integrates seamlessly with existing T2I models, using large language models to interpret and decompose complex prompts, thereby guiding the generation process more effectively. 
As existing evaluation metrics, such as CLIPScore, fail to adequately capture the nuances of complex instructions, we introduce an evaluation toolkit that automates the quality assessment of generated images using a set of multi-dimensional metrics. The data and code are released at \url{https://github.com/yczhou001/LongBench-T2I}.
\end{abstract}


\begin{teaserfigure}
  \vspace{-4mm}
  \includegraphics[width=\textwidth]{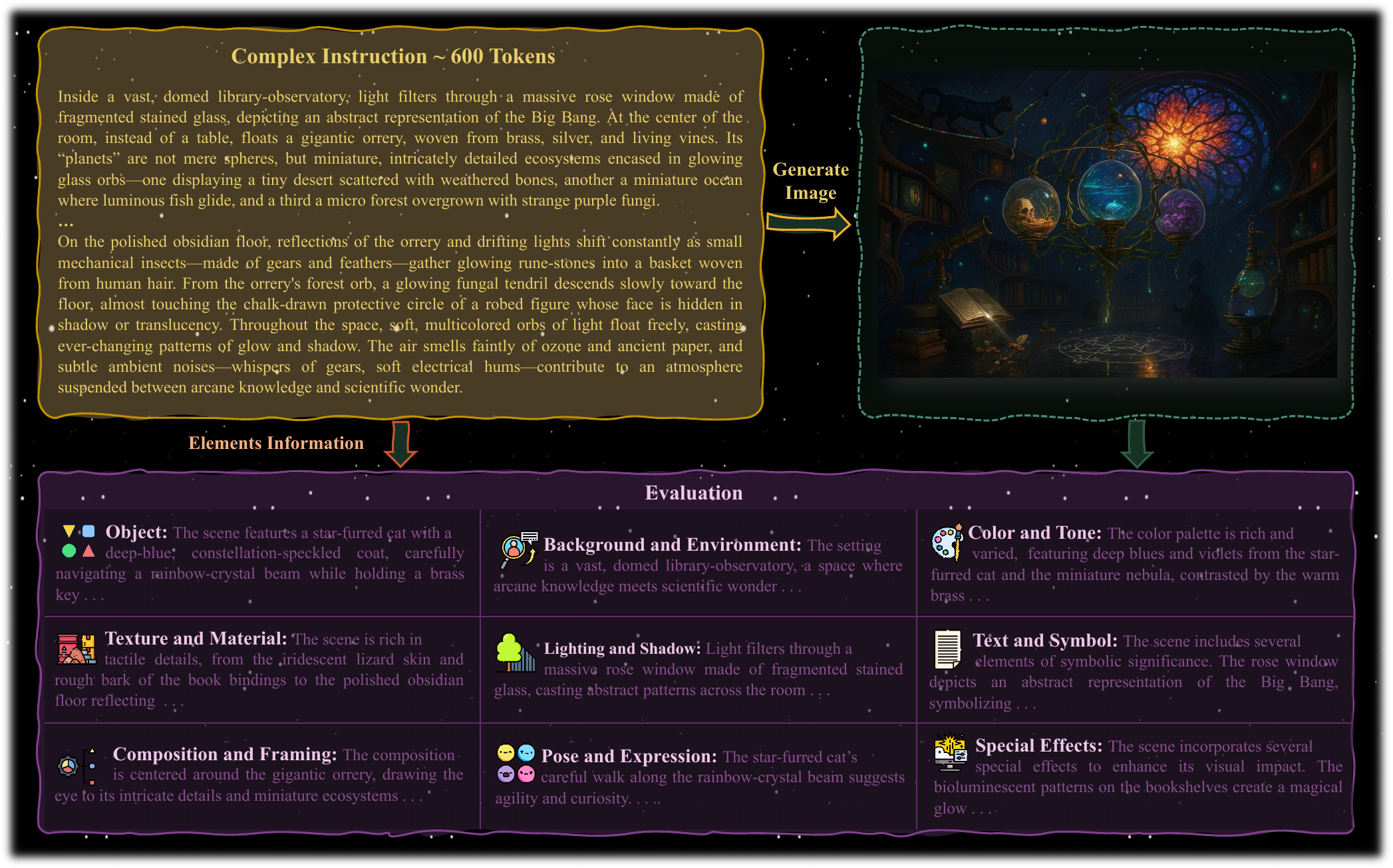}
  \vspace{-8mm}
  \caption{\small LongBench-T2I benchmark evaluates text-to-image models on their ability to follow complex instructions. The framework extracts nine key elements from each instruction to systematically assess the detail of the generated image.}
  \label{fig:teaser}
\end{teaserfigure}


\maketitle

\section{Introduction}
Recent advancements in text-to-image (T2I) generation have led to the development of models capable of producing high-quality images from textual descriptions. 
Many T2I models, e.g., DALL·E \cite{betker2023improving}, Stable Diffusion \cite{rombach2022high}, and Imagen  \cite{saharia2022photorealistic}, have demonstrated impressive abilities in generating diverse and visually appealing images. 
However, their performance often diminishes when tasked with generating images from complex, multi-faceted instructions that involve intricate object compositions, specific attributes, and detailed spatial relationships.

Existing benchmarks, such as DrawBench~\cite{otani2023toward}, DPG-Bench~\cite{hu2024ella}, and T2I-CompBench \cite{huang2023t2i}, have made significant strides in evaluating compositional capabilities by introducing categories like attribute binding and object relationships. 
While these benchmarks provide valuable insights, they primarily focus on individual aspects of compositionality and often lack comprehensive coverage of the multifaceted nature of complex instructions. 
For instance, T2I-CompBench \cite{huang2023t2i} categorizes prompts into attribute binding and object relationships but does not extensively address the nuanced interplay between these elements in complex scenes. 
Additionally, the evaluation metrics employed, such as CLIPScore \cite{hessel2021clipscore}, may not fully capture the intricacies of compositional understanding, as they often rely on image-text similarity measures that can be ambiguous in the context of complex instructions.

Furthermore, while some benchmarks have introduced evaluation metrics tailored to specific compositional challenges, there is a lack of standardized, multi-dimensional assessment tools that can holistically evaluate a model's ability to follow complex instructions \cite{huang2023t2i,trusca2024object,hu2024ella}. 
This gap in evaluation frameworks hinders the development of models that are not only capable of generating high-quality images but also adept at adhering to intricate and detailed instructions.

\begin{table}[t]\footnotesize
\centering
\caption{\small Comparison of other T2I benchmarks. 
}
\vspace{-3mm}
\label{tab:benchmark-comparison}
\begin{tabular}{lcccc}
\toprule
\textbf{Benchmark} & \textbf{Prompts} & \textbf{Source} & \textbf{Component} & \textbf{Long} \\
\midrule
DrawBench~\cite{otani2023toward} & 200 & Human & Object Level & 15  \\
T2I-CompBench~\cite{huang2023t2i} & 5,000 & Template & 3 Categories & 10 \\

DPG-Bench~\cite{hu2024ella} & 1,065 & Template & Object Level & 84 \\
\textbf{LongBench-T2I (Ours)} & \textbf{500} & LLM & \textbf{9 Elements} & \textbf{683} \\
\bottomrule
\end{tabular}
\vspace{-3mm}
\end{table}

To address these limitations, we propose \textbf{LongBench-T2I}, a comprehensive benchmark designed to evaluate T2I models under complex instructions. 
LongBench-T2I encompasses 500 meticulously curated prompts that span nine diverse visual evaluation dimensions, providing a robust framework to assess a model's instruction-following capabilities comprehensively. 
In addition to benchmarking, we introduce an agent framework, Plan2Gen, that facilitates complex instruction-based image generation without necessitating additional model training. 
This framework seamlessly integrates with existing T2I models, leveraging large language models (LLMs) to parse and decompose complex instructions, thereby guiding the image generation process more effectively.

By providing a holistic benchmark and a novel agent baseline, LongBench-T2I aims to bridge the existing gaps in the evaluation of complex instruction-following in T2I models. It can catalyze further research into developing more controllable and user-aligned generative models, advancing the field of complex instruction-based image generation.
The main contributions are summarized as follows:
\begin{itemize}[leftmargin=*]
\item We present LongBench-T2I, the first comprehensive benchmark for image generation under complex instructions, consisting of 500 intricate prompts covering 9 diverse visual evaluation dimensions, which enables a series of analyses on instruction-following capabilities of image generation models.
\item We conduct extensive pilot studies on various categories of text-to-image generation models using LongBench-T2I, providing comprehensive analyses across Closed-Source versus Open-Source models and Diffusion versus Autoregressive architectures.
\item We also propose Plan2Gen for complex instruction-based image generation, which requires no model training and can seamlessly integrate with existing image generation models under the direction of an LLM.
\item To facilitate future research, we release an evaluation package that automates result evaluation and supports multi-dimensional assessment, hoping LongBench-T2I will promote comprehensive studies on image generation from complex instructions and drive the development of more controllable and user-aligned models.
\end{itemize}

\section{Related Work}
\textbf{Text-to-Image Generation.}
Recent T2I models, including diffusion-based methods (e.g., Imagen \cite{saharia2022photorealistic}, Stable Diffusion \cite{rombach2022high}, FLUX \cite{flux2024}), autoregressive approaches (e.g., DALL·E \cite{betker2023improving}, Infinity \cite{han2024infinity}), and hybrid variants, have made notable progress in visual fidelity and semantic alignment \cite{liao2025step,zhou2024less,xu2023imagereward,dong2023raft,wallace2024diffusion,wu2024deep}. However, they often fail under complex prompts involving multiple entities, fine-grained attributes, and spatial relations. 
To improve controllability, recent works have introduced agent-based frameworks \cite{fang2025got,wang2024genartist,hahn2024proactive,qin2024diffusiongpt} and chain-of-thought \cite{guo2025can,jiang2025t2i} prompting to guide generation via LLM reasoning \cite{zhuang2025vargpt,zhou2024visual,zhou2025weak,wang2025mint} with self-correction \cite{wu2024self,fang2025got}, yet these methods remain limited in handling long-form compositionality due to the lack of structured generation and iterative validation. 

\begin{figure*}[t]
    \centering
    \includegraphics[width=1\linewidth]{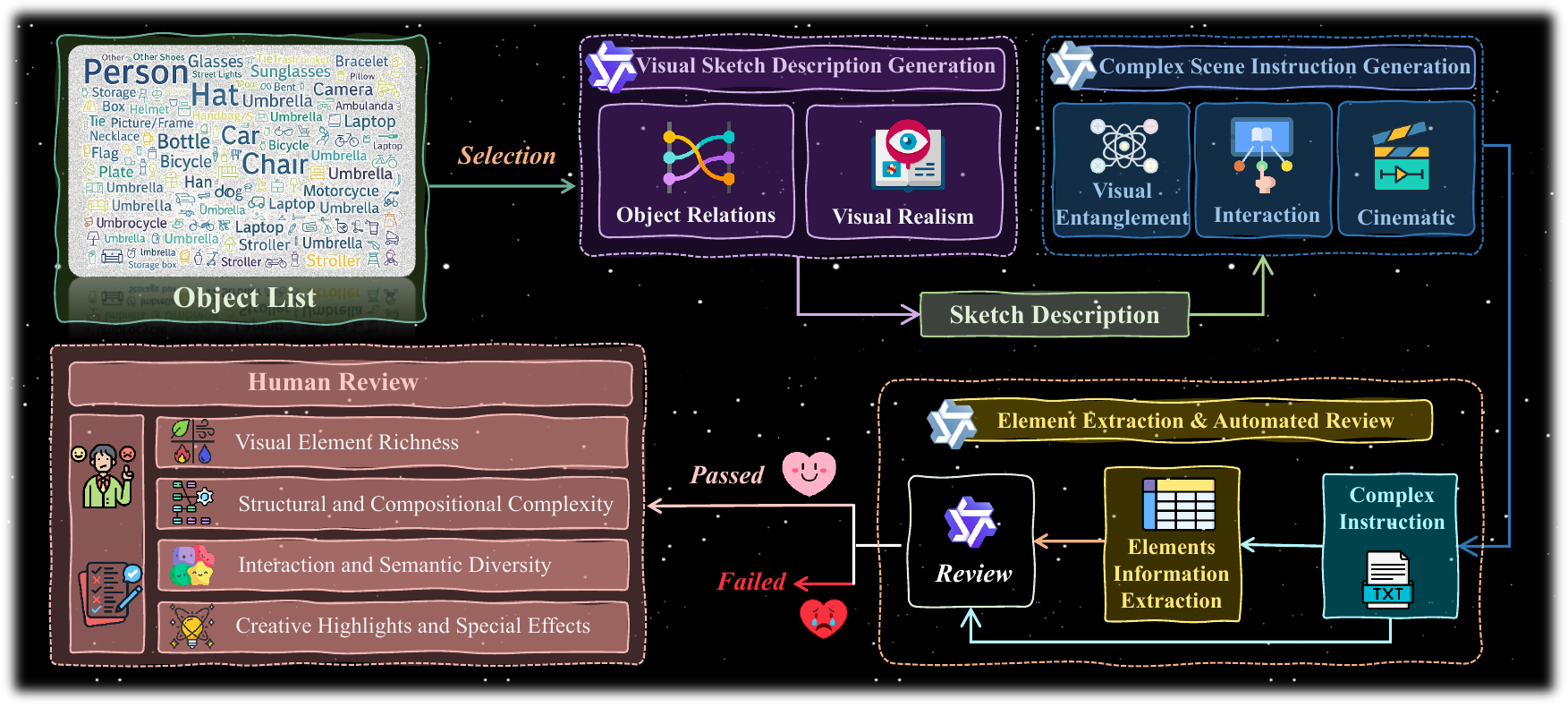}
    \vspace{-8mm}
    \caption{\small The multi-stage pipeline for generating the LongBench-T2I benchmark.}
    \label{fig:pipeline}
    \vspace{-3mm}
\end{figure*}

\noindent\textbf{Image Generation Benchmark.}
Existing benchmarks have been introduced to assess compositional text-to-image generation, yet each remains limited in scope. DrawBench \cite{otani2023toward}, with 200 short prompts, targets basic skills like counting and attribute binding, but relies on single-sentence inputs with limited semantic depth. DAA-200 \cite{trusca2024object} offers 200 adversarial prompts to test attribute binding precision, though its scope remains confined to simple two-object scenes. T2I-CompBench \cite{huang2023t2i} focuses on open-world compositionality with 5K prompts, but most are short and cover only isolated skills per instance. DPG-Bench \cite{hu2024ella} expands compositional density through template-based multi-object prompts, yet its auto-generated language lacks the descriptive nuance and narrative structure required for evaluating long-form instruction following. However, these are limited by brevity, narrow diversity, or synthetic prompts.
In contrast, as summarized in Table~\ref{tab:benchmark-comparison}, LongBench-T2I provides 500 multi-sentence prompts generated by LLMs and filtered by humans, each covering nine visual aspects for evaluating models under complex instructions.

\section{LongBench-T2I Benchmark}
\label{sec:benchmark_construction}
The LongBench-T2I benchmark is systematically developed through a multi-stage pipeline aimed at producing complex and high-quality text-to-image instructions. As illustrated in Figure~\ref{fig:pipeline}, the process utilizes LLMs for text generation and refinement, supported by meticulous human oversight to ensure the benchmark's quality.

\subsection{Sketch Description Generation}
The process commences with the generation of a foundational visual sketch. We begin by randomly sampling a single object from a predefined list, derived from the Object365 dataset~\citep{shao2019objects365}. This seed object is then fed into an LLM. The LLM is tasked with creating a descriptive visual sketch that incorporates this primary object. To foster diversity and contextual richness from the outset, the LLM is specifically prompted to consider \textit{Object Relations} (how the seed object might interact with or be situated relative to other potential objects) and \textit{Visual Realism}. The aim is to produce a varied set of initial descriptions that include multiple objects and lay the groundwork for more complex scenes, moving beyond simplistic single-object depictions.

\subsection{Complex Scene Instruction Generation}
Building upon the visual sketch obtained in the previous stage, we elaborate this sketch into a comprehensive, intricate, and detailed textual instruction suitable for advanced text-to-image models. This employs an LLM to enhance the description by incorporating considerations of \textit{visual entanglement} (e.g., objects occluding or intertwining with each other), complex \textit{interactions} between multiple entities, and \textit{cinematic qualities} (such as specific camera angles, depth of field, or dramatic lighting). The objective of this stage is to substantially increase the complexity and narrative depth of the textual prompt, pushing the boundaries of what current T2I models can interpret and render.

\begin{figure*}[!t]
    \centering
    \includegraphics[width=1\linewidth]{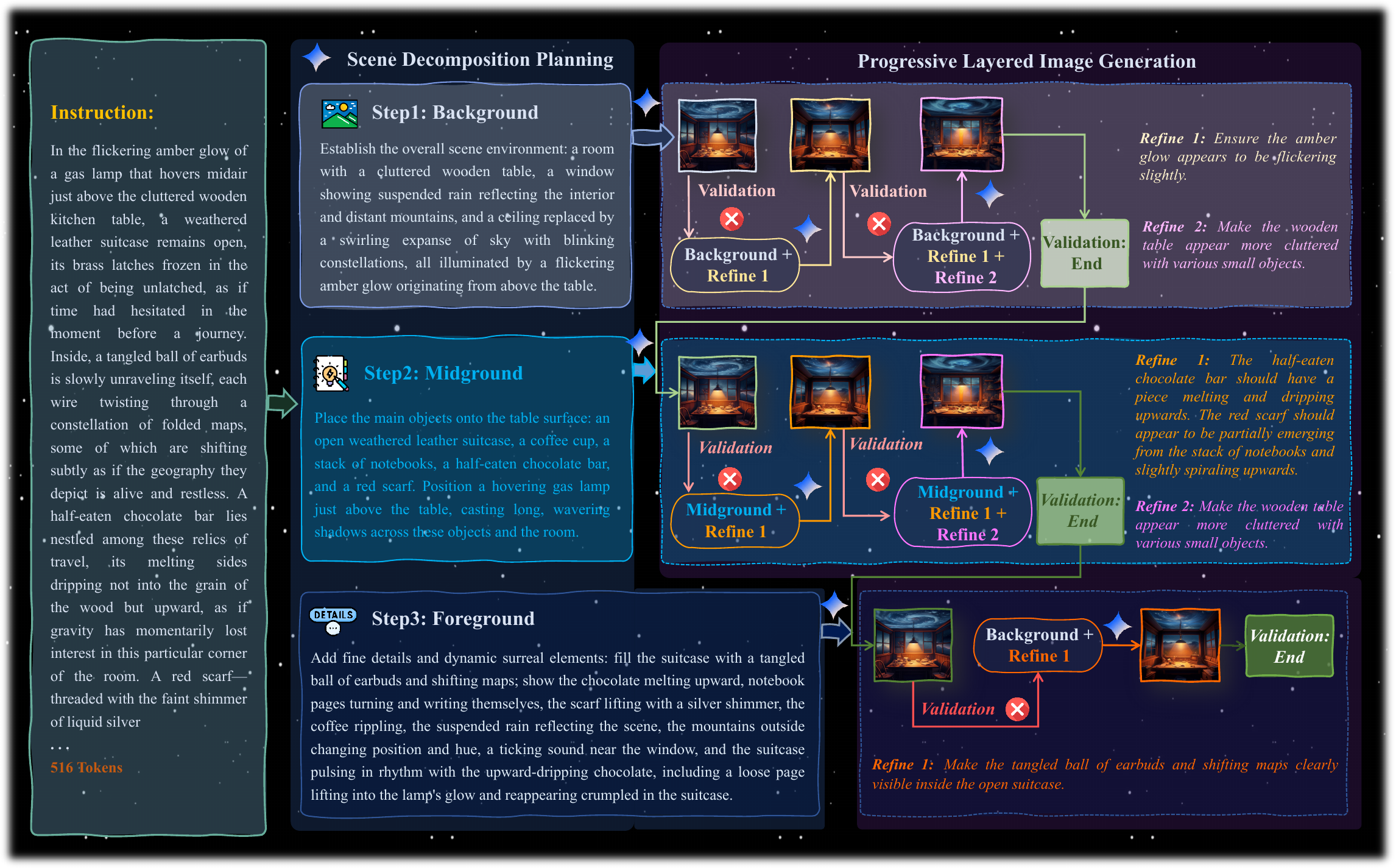}
    \vspace{-8mm}
    \caption{\small The Plan2Gen Agent generates images from complex long instructions by first decomposing the scene into background, midground, and foreground prompts, then progressively generating and refining each layer with validation to ensure alignment with the original description.}
    \label{fig:agent}
    \vspace{-3mm}
\end{figure*}

\subsection{Element Extraction and Automated Review}\label{sec:element}
To ensure the structural integrity and semantic coherence of the generated complex scene instructions, a subsequent stage implements element extraction and automated review. An LLM is utilized to parse each complex instruction and extract information on nine predefined key visual elements:
\ding{182} \textit{Object(s)}: Main subjects and supporting items. 
\ding{183} \textit{Background and Environment}: Setting and surrounding context. 
\ding{184} \textit{Color and Tone}: Dominant hues, mood, and atmosphere.
\ding{185} \textit{Texture and Material}: Surface properties of objects.
\ding{186} \textit{Lighting and Shadow}: Illumination sources, direction, and resulting shadows.
\ding{187} \textit{Text and Symbol}: Any textual or symbolic elements present.
\ding{188} \textit{Composition and Framing}: Arrangement of elements and camera perspective.
\ding{189} \textit{Pose and Expression}: Postures and facial expressions of animate subjects.
\ding{190} \textit{Special Effects}: Any unique visual augmentations (e.g., motion blur, lens flare).
Following extraction, the LLM performs an automated validation. It cross-references the extracted elements against the original instruction to check for consistency and completeness.

\subsection{Human Review}
The concluding stage involves a rigorous human review process for all instructions that have successfully passed the LLM-based automated validation. Two graduate students meticulously evaluate each candidate's instruction based on four crucial dimensions:
\ding{182} \textit{Visual Element Richness}: The diversity, detail, and specificity of the described visual components.
\ding{183} \textit{Structural and Compositional Complexity}: The intricacy of object arrangements, scene layout, and spatial relationships.
\ding{184} \textit{Interaction and Semantic Diversity}: The presence of meaningful interactions between elements and the overall semantic coherence and novelty of the scene.
\ding{185} \textit{Creative Highlights and Special Effects}: The inclusion of imaginative concepts, unique artistic styles, or challenging visual effects that test advanced generation capabilities.
Only instructions meeting these criteria are included in LongBench-T2I, ensuring the benchmark’s quality and relevance for evaluating long-context text-to-image generation.

\section{Plan2Gen Agent}
\label{sec:plan2gen_agent}
The Plan2Gen agent tackles complex, long-form image generation through a structured, multi-stage approach (Figure~\ref{fig:agent}), with scene planning, iterative generation, and validation.

\subsection{Scene Decomposition Planning}
Given a complex and lengthy textual instruction, an LLM is first tasked with a strategic planning role. The LLM analyzes the entirety of the input text to identify and segregate the described visual elements into distinct compositional layers. Specifically, it deconstructs the overall scene into three core components:
\ding{182} \textbf{Background}: Elements constituting the furthest plane of the scene, establishing the overall environment and atmosphere (e.g., sky, distant mountains, room walls).
\ding{183} \textbf{Midground}: Objects and entities situated between the background and foreground, often serving as primary or secondary subjects (e.g., a table, a suitcase, a coffee cup upon the table).
\ding{184} \textbf{Foreground}: Elements closest to the viewer, or those requiring fine-grained detail and prominence, potentially interacting with or overlaying midground elements (e.g., specific items within an open suitcase, detailed textures, immediate atmospheric effects).
This decomposition yields three layer-specific sub-prompts, enabling the model to focus on manageable details at each stage and progressively build scene complexity.

\subsection{Progressive Layered Image Generation}
This process constructs the final image by iteratively generating and refining each compositional layer, background, midground, and foreground, in sequence. For each layer, a consistent generation and refinement pipeline is applied:
\ding{182} \textbf{Initial Generation:} The layer-specific sub-prompt (derived from the decomposition phase) is input to an LLM with image generation capabilities (or a specialized text-to-image model). For midground and foreground layers, the generation is also conditioned on the successfully generated preceding layer(s) to ensure contextual coherence.
\ding{183} \textbf{Validation:} The generated image for the current layer is assessed by an LLM against the requirements outlined in its corresponding sub-prompt.
\ding{184} \textbf{Iterative Refinement:} If the validation indicates a mismatch with the prompt (e.g., the ``amber glow'' not ``flickering slightly'' for the background, or issues with the ``half-eaten chocolate bar'' for the midground, as depicted in Figure~\ref{fig:agent}), the original sub-prompt for that layer is augmented with specific refinement instructions. This revised prompt is then used to re-generate the image for the current layer. This generation-validation-refinement cycle can be repeated.
\ding{185} \textbf{Progression Criterion:} The refinement loop for the current layer terminates when either the generated image successfully passes validation or a predefined maximum number of refinement attempts is exhausted.

This pipeline is applied sequentially to the background, midground, and foreground layers, with each stage conditioned on the approved outputs of previous layers. Layer-by-layer generation, validation, and refinement help Plan2Gen manage complexity and improve image fidelity to long-form instructions. The maximum refinement threshold ensures efficient termination.

\section{Experiments}
\subsection{Experimental Setup}
\textbf{Models}
To comprehensively assess the capabilities of current text-to-image generation systems on long-context instructions, we evaluate a diverse range of models using our LongBench-T2I benchmark. The evaluated models span leading closed-source systems, prominent open-source alternatives, and architectures including both autoregressive and diffusion-based approaches. This broad selection allows for a thorough comparison and highlights the challenges posed by our benchmark across different model types, including the performance of our proposed Plan2Gen agent.

\noindent\textbf{Evaluation Metrics}
To quantitatively and qualitatively assess image fidelity to the LongBench-T2I instructions, we evaluate generated images across the nine key visual elements defined in Section~\ref{sec:element}: Object(s) (\textbf{Obj.}), Background and Environment (\textbf{Backg.}), Color and Tone (\textbf{Color}), Texture and Material (\textbf{Texture}), Lighting and Shadow (\textbf{Light}), Text and Symbol (\textbf{Text}), Composition and Framing (\textbf{Comp.}), Pose and Expression (\textbf{Pose}), and Special Effects (\textbf{FX}).
Evaluation utilizes MLLMs: Google's Gemini-2.0-Flash (closed-source) and InternVL3-78B (open-source).
These MLLMs assess generated images against nine visual elements specified in input instructions.
The open-source InternVL3-78B is employed to ensure long-term reproducibility of the evaluation results.

\begin{table}[t]\small
\centering
\caption{\small Dimension-wise average scores for all models across different evaluation dimensions, evaluated by \textbf{Gemini-2.0-Flash}.}
\label{tab:main_results}
\vspace{-3mm}
\setlength{\tabcolsep}{1.5pt}
\resizebox{\linewidth}{!}{
\begin{tabular}{lcccccccccc}
\toprule
\bf Method & \textbf{Obj.} & \textbf{Backg.} & \textbf{Color} & \textbf{Texture} & \textbf{Light} & \textbf{Text} & \textbf{Comp.} & \textbf{Pose} & \textbf{FX} & \textbf{Avg.} \\
\midrule
\multicolumn{11}{c}{\textit{Diffusion-based Methods}} \\
\midrule
Nexus-Gen \cite{zhang2025nexus} & 1.71 & 3.19 & 3.27 & 3.34 & 2.27 & 1.35 & 2.44 & 1.81 & 1.38 & 2.31 \\
GoT \cite{fang2025got} & 2.50 & 3.10 & 3.84 & 3.50 & 2.66 & 1.80 & 2.97 & 2.10 & 1.99 & 2.72 \\
FLUX.1-dev \cite{flux2024} & 3.24 & 3.46 & 4.04 & 3.91 & 3.29 & 2.11 & 3.78 & 2.71 & 1.72 & 3.14 \\
Omnigen \cite{xiao2024omnigen} & 3.14 & 3.70 & 4.18 & 3.71 & 3.02 & 2.42 & 3.81 & 2.69 & 2.55 & 3.25 \\
\midrule
\multicolumn{11}{c}{\textit{AR-based Methods}} \\
\midrule
LlamaGen-3B \cite{sun2024autoregressive} & 1.01 & 1.34 & 2.35 & 2.05 & 1.66 & 1.35 & 1.03 & 1.12 & 1.85 & 1.53 \\
Janus-pro-1B \cite{chen2025janus} & 2.23 & 2.60 & 2.94 & 2.92 & 2.09 & 1.58 & 2.36 & 1.84 & 1.60 & 2.24 \\
Janus-pro-7B \cite{chen2025janus} & 2.67 & 2.95 & 3.54 & 3.34 & 2.61 & 1.74 & 2.83 & 2.01 & 1.73 & 2.60 \\
Infinity-2B \cite{han2024infinity} & 2.99 & 3.49 & 4.09 & 3.65 & 2.96 & 2.14 & 3.54 & 2.59 & 2.69 & 3.13 \\
Infinity-2B-reg \cite{han2024infinity} & 3.05 & 3.62 & 4.13 & 3.85 & 3.21 & 2.14 & 3.54 & 2.44 & 2.13 & 3.12 \\
Infinity-8B \cite{han2024infinity} & 3.36 & 3.96 & 4.34 & 4.12 & 3.54 & 2.48 & 4.08 & 2.72 & 2.35 & 3.44 \\
Lumina-mGPT \cite{liu2024lumina} & 3.20 & 3.38 & 4.10 & 4.02 & 3.24 & 2.14 & 3.47 & 2.56 & 1.89 & 3.11 \\
Gemini-2.0 \cite{gemini2025flash} & 3.60 & 3.75 & 4.34 & 3.91 & 3.39 & 2.64 & 4.21 & 3.02 & 2.56 & 3.49 \\
Gemini-2.0 w/ CoT & 2.79 & 3.20 & 3.91 & 3.37 & 3.20 & 2.16 & 3.26 & 2.47 & 2.34 & 2.96 \\
GPT-4o \cite{hurst2024gpt} & \textbf{3.94} & 4.03 & \textbf{4.56} & \textbf{4.17} & 3.39 & 2.86 & 4.22 & \textbf{3.36} & 2.78 & 3.70 \\
\midrule
\textbf{Plan2Gen (Ours)} & 3.76 & \textbf{4.22} & 4.43 & 3.80 & \textbf{3.72} & \textbf{2.86} & \textbf{4.43} & 3.29 & \textbf{3.06} & \textbf{3.73} \\
\bottomrule
\end{tabular}
}
\vspace{-3mm}
\end{table}

\begin{table}[t]\small
\centering
\caption{\small Dimension-wise average scores for all models, evaluated by \textbf{InternVL3-78B} \cite{zhu2025internvl3}.}
\label{tab:internvl_results}
\vspace{-3mm}
\setlength{\tabcolsep}{1.5pt}
\resizebox{\linewidth}{!}{
\begin{tabular}{lcccccccccc}
\toprule
\bf Method & \textbf{Obj.} & \textbf{Backg.} & \textbf{Color} & \textbf{Texture} & \textbf{Light} & \textbf{Text} & \textbf{Comp.} & \textbf{Pose} & \textbf{FX} & \textbf{Avg.} \\
\midrule
\multicolumn{11}{c}{\textit{Diffusion-based Methods}} \\
\midrule
Nexus-Gen \cite{zhang2025nexus}  & 1.78 & 2.88 & 2.84 & 2.90 & 2.42 & 1.31 & 2.35 & 1.64 & 1.44 & 2.17 \\
GoT \cite{fang2025got} & 2.28 & 2.91 & 3.33 & 3.07 & 2.63 & 1.67 & 2.84 & 1.91 & 1.99 & 2.51 \\
FLUX.1-dev \cite{flux2024} & 2.86 & 3.04 & 3.52 & 3.39 & 2.99 & 1.92 & 3.47 & 2.26 & 1.55 & 2.78 \\
Omnigen \cite{xiao2024omnigen}  & 2.79 & 3.25 & 3.67 & 3.37 & 2.84 & 2.29 & 3.48 & 2.41 & 2.56 & 2.96 \\
\midrule
\multicolumn{11}{c}{\textit{AR-based Methods}} \\
\midrule
LlamaGen-3B \cite{sun2024autoregressive} & 1.00 & 1.08 & 1.57 & 1.35 & 1.14 & 1.01 & 1.01 & 1.02 & 1.56 & 1.19 \\
Janus-pro-1B \cite{chen2025janus} & 2.16 & 2.61 & 2.70 & 2.59 & 2.27 & 1.56 & 2.37 & 1.88 & 1.81 & 2.21 \\
Janus-pro-7B \cite{chen2025janus} & 2.47 & 2.91 & 3.15 & 3.01 & 2.66 & 1.69 & 2.83 & 1.97 & 1.85 & 2.50 \\
Infinity-2B \cite{han2024infinity} & 2.72 & 3.38 & 3.82 & 3.52 & 3.03 & 2.10 & 3.44 & 2.24 & 2.49 & 2.97 \\
Infinity-2B-reg~\cite{han2024infinity} & 1.78 & 2.88 & 2.84 & 2.90 & 2.42 & 1.31 & 2.35 & 1.64 & 1.44 & 2.17 \\
Infinity-8B \cite{han2024infinity} & 2.98 & 3.53 & 3.97 & 3.73 & 3.30 & 2.30 & 3.79 & 2.36 & 2.21 & 3.13 \\
Lumina-mGPT \cite{liu2024lumina} & 2.80 & 3.10 & 3.57 & 3.40 & 3.00 & 1.96 & 3.18 & 2.23 & 1.79 & 2.78 \\

Gemini-2.0 \cite{gemini2025flash} & 3.10 & 3.29 & 3.96 & 3.48 & 3.15 & 2.34 & 3.88 & 2.66 & 2.51 & 3.15 \\
Gemini-2.0 w/ CoT & 2.47 & 2.94 & 3.46 & 3.03 & 2.90 & 2.00 & 2.97 & 2.12 & 2.43 & 2.70 \\
GPT-4o \cite{hurst2024gpt} & \textbf{3.33} & 3.53 & \textbf{4.27} & \textbf{4.13} & 3.47 & 2.30 & 3.93 & \textbf{2.93} & 2.60 & 3.39 \\\midrule
\textbf{Plan2Gen (Ours)} & 3.28 & \textbf{3.76} & 4.09 & 3.53 & \textbf{3.51} & \textbf{2.56} & \textbf{4.18} & 2.92 & \textbf{2.89} & \textbf{3.41} \\
\bottomrule
\end{tabular}
}
\vspace{-3mm}
\end{table}

\subsection{Performance Comparison}
We evaluated Plan2Gen against various text-to-image models on LongBench-T2I, using Gemini-2.0-Flash (Table~\ref{tab:main_results}) and InternVL3-78B (Table~\ref{tab:internvl_results}) as evaluators.
(1) \textbf{Overall Performance:}  
Plan2Gen achieves the highest average scores (3.73 on Gemini-2.0-Flash, 3.41 on InternVL3-78B), slightly outperforming the strong proprietary model GPT-4o. This demonstrates Plan2Gen's effectiveness on complex instructions.
(2) \textbf{Closed- and Open-Source Alternatives:}  
Plan2Gen, leveraging Gemini-2.0, achieves state-of-the-art results, rivaling or surpassing GPT-4o. Among open-source models, the Infinity series (AR-based) and Omnigen (diffusion-based) also perform well, but Plan2Gen’s planning mechanism offers an advantage on long, complex prompts.
(3) \textbf{Diffusion vs. AR Architectures:}  
The results, grouped by ``Diffusion-based'' and ``AR-based'' methods, show that AR-based models (Plan2Gen, GPT-4o) achieve top scores, especially with advanced planning. While some AR models lag behind leading diffusion models, Plan2Gen’s layered approach excels in maintaining coherence on lengthy, multi-faceted prompts.
(4) \textbf{Dimensional Strengths:}  
Plan2Gen leads in ``Background'', ``Lighting'', ``Text'', ``Composition'', and ``Special Effects (FX)''. GPT-4o performs best on ``Object(s)'', ``Color'', ``Texture'', and ``Pose'', though Plan2Gen remains highly competitive. Notably, both Plan2Gen and GPT-4o achieve the best results on the challenging ``Text'' dimension.
(5) \textbf{Evaluator Consistency:}  
Model rankings and performance trends are consistent across both evaluators, with Plan2Gen and GPT-4o as top performers. This cross-evaluator agreement further validates our findings.

\begin{figure}[t]
    \centering
    \includegraphics[width=\linewidth]{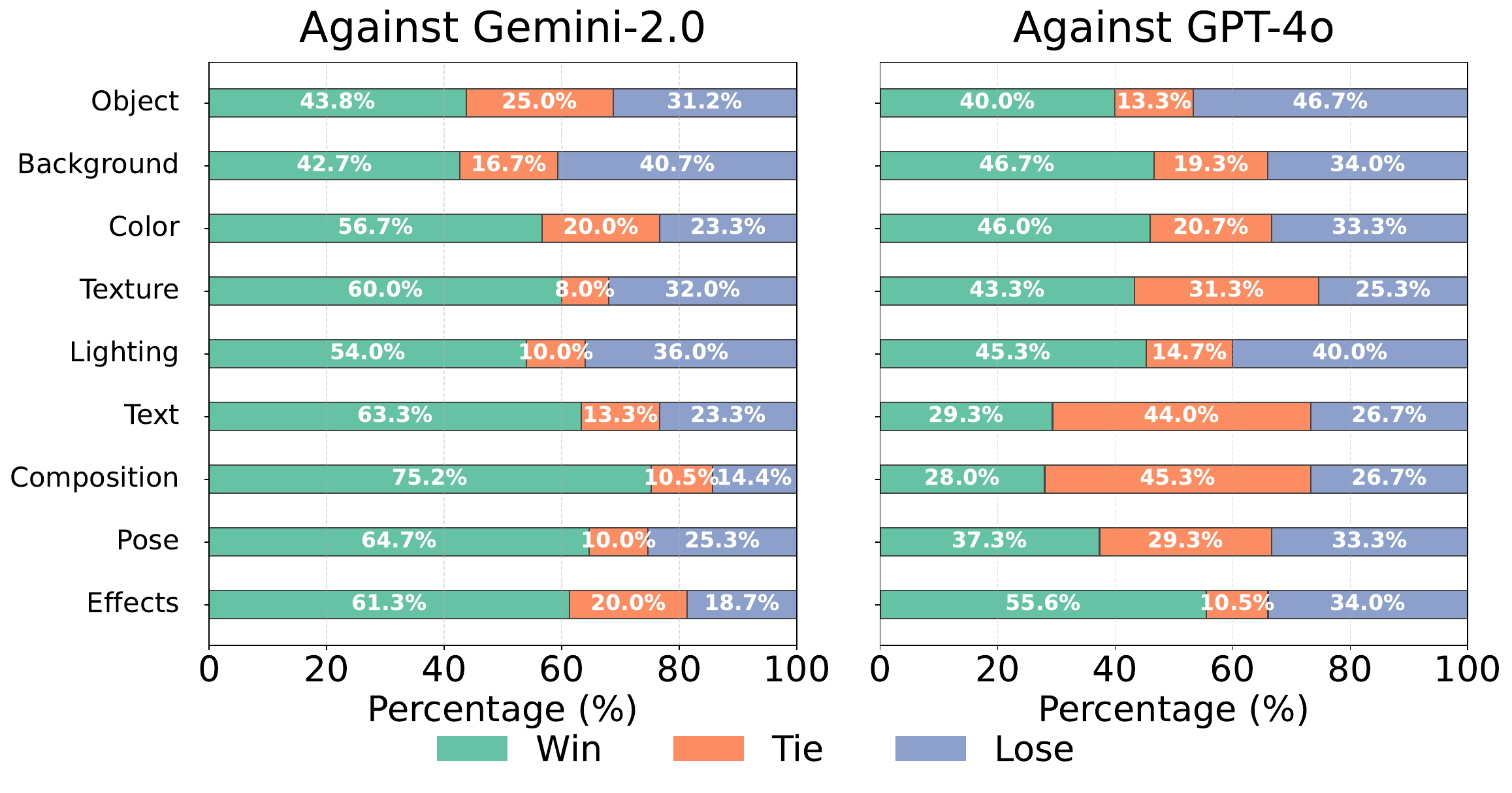}
    \vspace{-7mm}
    \caption{\small Comparative performance evaluation across nine visual dimensions when our model is compared against Gemini-2.0 (left) and GPT-4o (right). Each horizontal bar shows the proportion of images where our model was rated as better (Win), equivalent (Tie), or worse (Lose).}
    \label{fig:human}
    \vspace{-3mm}
\end{figure}
\subsection{Human Evaluation}
\label{sec:human_evaluation}
To complement our automated metrics, we conducted pairwise human evaluations comparing Plan2Gen against Gemini-2.0 and GPT-4o on a subset of benchmark prompts. Evaluators chose the image that better represented the prompt across nine visual dimensions, with results shown in Figure~\ref{fig:human}. Against Gemini-2.0 (Figure~\ref{fig:human}, left), Plan2Gen demonstrated a significant advantage, achieving higher win rates in all nine dimensions. The comparison with GPT-4o (Figure~\ref{fig:human}, right) was more competitive. Plan2Gen achieved a higher win rate than loss rate in eight of the nine dimensions. GPT-4o held a slight edge only in ``Object'' fidelity.

\begin{figure}[t]
    \centering
    \includegraphics[width=0.59\linewidth]{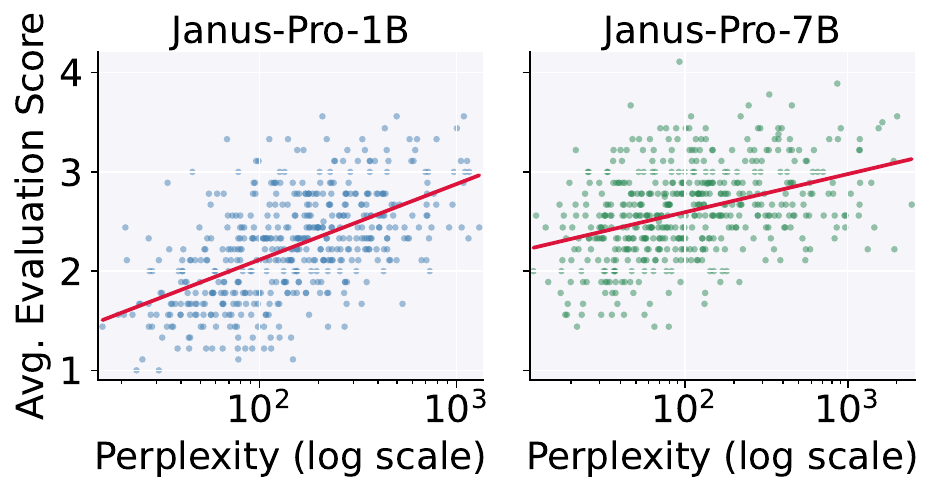}
    \includegraphics[width=0.395\linewidth]{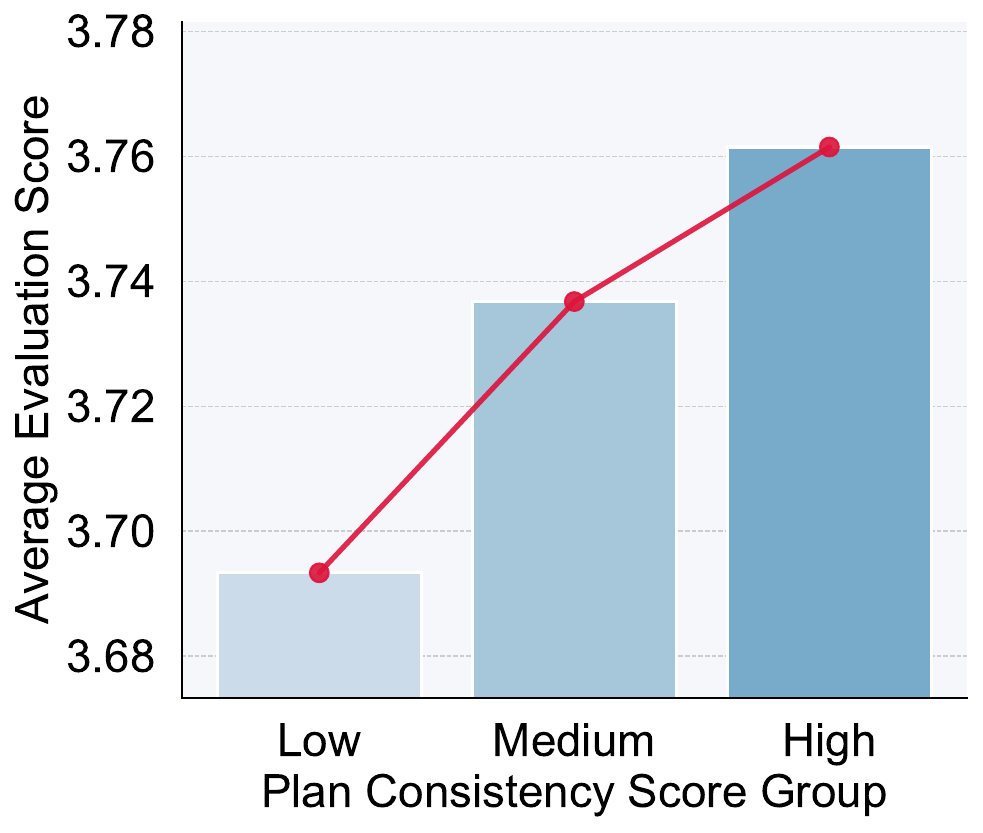}
    \vspace{-7mm}
    \caption{\small (Left) Instruction perplexity vs. average evaluation score for Janus-Pro models. Each point represents a sample. The red line indicates the linear fit in $\log_{10}$ perplexity space.
    (Right) Relationship Between Plan Consistency and Generation Quality.}
    \label{fig:perplexity_vs_score}
    \vspace{-3mm}
\end{figure}
\subsection{Empowering Visual Generation with Language Understanding?}
\label{sec:empowering_visual_gen}
We investigated the relationship between a model's language understanding, gauged by instruction perplexity (PPL), and the quality of its visual outputs. Figure~\ref{fig:perplexity_vs_score} (left) reveals a notable trend for the Janus-Pro-1B model: a counter-intuitive positive correlation where prompts with higher PPL (indicating poorer textual comprehension by the model) sometimes yielded images with higher evaluation scores. This suggests a discernible gap between the current language understanding capabilities of such multi-modal models and their ability to translate that understanding into high-fidelity visual generation; simply achieving low PPL on a prompt does not guarantee superior image output. Interestingly, this counter-intuitive trend is notably attenuated in the larger Janus-Pro-7B model. The general improvement in scores and a less pronounced positive PPL-score correlation for the 7B model suggest that increasing model scale can, to some extent, help bridge this gap, leading to a more consistent (though still imperfect) alignment where better text understanding more reliably yields higher-quality images. This highlights that while language understanding is a target, current MLLMs, especially smaller ones, find it particularly challenging to effectively translate their language understanding into visual synthesis, a difficulty that larger ones appear to partially alleviate.

\subsection{Analysis of Plan2Gen}

\begin{table}[t]\small
\centering
\caption{\small \small
Dimension-wise scores across different refinement steps. 
}
\label{tab:refine_scores}
\vspace{-3mm}
\setlength{\tabcolsep}{1.5pt}
\resizebox{\linewidth}{!}{
\begin{tabular}{ccccccccccc}
\toprule
\textbf{Steps} & \textbf{Obj.} & \textbf{Backg.} & \textbf{Color} & \textbf{Texture} & \textbf{Light} & \textbf{Text} & \textbf{Comp.} & \textbf{Pose} & \textbf{FX} & \textbf{Avg.} \\
\midrule
1 & 3.76 & 4.17 & 4.40 & 3.77 & 3.67 & 2.79 & 4.38 & 3.21 & 2.95 & 3.68 \\
3 & 3.76 & 4.22 & 4.43 & 3.80 & 3.72 & 2.86 & 4.43 & 3.29 & 3.06 & \bf 3.73 \\
5 & 3.76 & 4.11 & 4.47 & 3.77 & 3.71 & 2.79 & 4.45 & 3.30 & 3.05 & 3.71 \\
7 & 3.70 & 4.12 & 4.42 & 3.74 & 3.65 & 2.75 & 4.29 & 3.18 & 3.02 & 3.65 \\
\bottomrule
\end{tabular}
}
\vspace{-3mm}
\end{table}
\textbf{Impact of Refine Iteration}
We investigated the effect of varying the maximum number of refinement iterations per layer within the Plan2Gen agent, with results shown in Table~\ref{tab:refine_scores}. The findings demonstrate that iterative refinement is beneficial, as performance improves when increasing from a single refinement step. The overall average score peaks when allowing up to 3 refinement steps. Beyond this point, allowing more iterations (e.g., 5 or 7 steps) does not consistently improve results and can lead to a slight degradation in overall average scores. This suggests that while a few refinement cycles are effective for enhancing alignment with sub-prompts, excessive iterations may risk over-correction or semantic drift. Therefore, a maximum of 3 refinement steps per layer offers an optimal balance, and this configuration was used for Plan2Gen in our main experiments.

\begin{table}[t]\small
\centering
\caption{\small \small
Comparison of different planning models.}
\label{tab:planning-results}
\vspace{-3mm}
\setlength{\tabcolsep}{1.5pt}
\resizebox{\linewidth}{!}{
\begin{tabular}{lcccccccccc}
\toprule
\textbf{Planner} & \textbf{Obj.} & \textbf{Backg.} & \textbf{Color} & \textbf{Texture} & \textbf{Light} & \textbf{Text} & \textbf{Comp.} & \textbf{Pose} & \textbf{FX} & \textbf{Avg.} \\
\midrule
Plan2Gen (Gemini 2.0 Flash) & \textbf{3.76} & \textbf{4.22} & \textbf{4.43} & \textbf{3.80} & \textbf{3.72} & \textbf{2.86} & \textbf{4.43} & \textbf{3.29} & \textbf{3.06} & \textbf{3.73} \\
Plan2Gen (Gemini 1.5 Flash-8b)      & 3.41 & 3.63 & 4.26 & 3.54 & 3.55 & 2.50 & 3.89 & 3.25 & 2.76 & 3.42 \\
\bottomrule
\end{tabular}
}
\vspace{-3mm}
\end{table}
\noindent\textbf{Planning Analysis}
The initial scene decomposition plan significantly impacts Plan2Gen's performance. Table~\ref{tab:planning-results} shows that using a more capable LLM (Gemini 2.0 Flash) as the planner leads to markedly better final image quality compared to a less capable one (Gemini 1.5 Flash-8b), underscoring the importance of high-quality planning. Furthermore, we assessed plan quality by evaluating the consistency of decomposed sub-prompts against the original instruction using Qwen3-32B, categorizing plans into Low, Medium, and High consistency groups. Figure~\ref{fig:perplexity_vs_score} (right) clearly demonstrates a positive correlation: higher plan consistency scores directly correspond to higher average evaluation scores for the final generated images. These findings validate that a well-structured and faithful plan is a key determinant for successfully generating complex scenes from long-form descriptions.

\section{Conclusion}
\label{sec:conclusion}
In this work, we introduced LongBench-T2I, a novel benchmark with complex, high-quality instructions for evaluating long-context text-to-image generation, and Plan2Gen, an agent framework that tackles these instructions via strategic scene decomposition and progressive layered refinement. Our experiments demonstrate that Plan2Gen achieves state-of-the-art performance, outperforming diverse models on LongBench-T2I, with human evaluations corroborating these results. LongBench-T2I and Plan2Gen serve as useful resources for the community in the ongoing development of image generation models for complex and detailed instructions.

\bibliographystyle{ACM-Reference-Format}
\bibliography{ref}
\end{document}